\documentclass[letterpaper, 10 pt, conference]{ieeeconf}  

\IEEEoverridecommandlockouts                              

\usepackage{graphics} 
\usepackage{epsfig} 
\usepackage{amsmath} 
\usepackage{amssymb}  
\usepackage{siunitx}
\usepackage{booktabs}

\title{\LARGE \bf
Towards a Physics Engine to Simulate
Robotic Laser Surgery: Finite Element Modeling
of Thermal Laser-Tissue Interactions
}

\author{Nicholas E. Pacheco, Kang Zhang, Ashley S. Reyes, Christopher J. Pacheco, Lucas Burstein, and Loris Fichera
\thanks{This material is based upon work supported by the National Science Foundation (NSF) under grant numbers 2237011 and 
2038257. Any opinions, findings, conclusions, or recommendations expressed in this material are those of the authors and do not necessarily reflect the views of the NSF.}
\thanks{N.E. Pacheco, K. Zhang, A.S. Reyes, C.J. Pacheco, L. Burstein, and L. Fichera are with the Department of Robotics Engineering,
       Worcester Polytechnic Institute, Worcester, MA 01690, USA.}%
\thanks{Corresponding author: Nicholas E. Pacheco (nepacheco@wpi.edu)}%
}

\begin{document}

\maketitle
\thispagestyle{empty}
\pagestyle{empty}

\begin{abstract}
This paper presents a computational model, based on the Finite Element Method (FEM), that simulates the thermal response of laser-irradiated tissue.
This model addresses a gap in the current ecosystem of surgical robot simulators, which generally lack support for lasers and other energy-based end effectors.
In the proposed model, the thermal dynamics of the tissue are calculated as the solution to a heat conduction problem with appropriate boundary conditions.
The FEM formulation allows the model to capture complex phenomena, such as convection, which is crucial for creating realistic simulations.
The accuracy of the model was verified via benchtop laser-tissue
interaction experiments using agar tissue phantoms and ex-vivo
chicken muscle. The results revealed an average root-mean-square
error (RMSE) of less than 2 \unit{\degreeCelsius} across most
experimental conditions.
\end{abstract}

\section{INTRODUCTION}
Computer simulations play a crucial role in surgical robotics
research, providing a safe and controlled environment 
where new robots and control algorithms can be tested
before they are used in actual surgeries.
With the growing interest in surgical robot
automation~\cite{Attanasio2021,Yang2017}, simulators
have become even more vital, offering a virtual space for
artificial intelligence agents to learn and practice surgical tasks.
Numerous open-source simulation frameworks 
for surgical robotics have been proposed in recent years~\cite{Schmidgall2024,Yu2024,Scheikl2023,Varier2022,Munawar2022,Xu2021,Tagliabue2020,Munawar2019,Fontanelli2018}.
These frameworks rely on third-party physics engines like SOFA~\cite{Faure2012}, Bullet~\cite{Coumans2021}, and PhysX\textregistered~\cite{PhysX} to render surgical tools and their interactions with human tissue.
While these frameworks can simulate many common surgical
instruments (e.g., scalpels, grippers, and needles), they
generally lack support for surgical lasers.
In surgery, lasers serve two main purposes, i.e., 
as cutting tools and for tissue
coagulation~\cite{Lee2022,Mattos2021,Fichera2021}.
Unlike scalpels and grippers, which use mechanical force to
cut or manipulate tissue, surgical lasers work contactlessly
and achieve their effect through heating~\cite{Lee2022}.
Unfortunately, the physics engines commonly used for robotic 
simulations do not support the thermal dynamics required to 
simulate laser-tissue interactions.
In this manuscript, we address this gap by proposing
a new computational model based on the Finite Element Method
(FEM) that accurately captures the thermal response of
laser-irradiated tissue.
Thermal laser-tissue interactions are generally considered
hard to model because they involve multiple complex 
physical processes, including light propagation
and heat transfer~\cite{Niemz2019}.
These interactions are influenced by numerous factors, including the laser's specific wavelength, power, and exposure time as well as the specific properties of the targeted tissue.
Previous work within the surgical robotics literature attempted
to model laser-tissue interactions using machine learning 
approaches~\cite{Pardo2015,Pardo2014}, which can be effective
but require the collection of extensive, high-quality datasets 
for training.
In more recent work~\cite{Arnold2022}, our group explored the use of
MCmatlab, an open-source library 
developed within the physics community~\cite{Marti2018}. 
MCmatlab provides accurate simulations of light propagation in tissue,
but its support for the simulation of thermal dynamics is limited,
particularly as it pertains to the handling of complex boundary conditions.
As we show in this manuscript, selecting appropriate boundary conditions is key to creating realistic simulations of surgical laser-tissue interactions.
This manuscript is organized as follows: Section II describes the
proposed FEM model for surgical laser-tissue interactions;
Section III reports on benchtop experiments 
performed on two types of tissue, showing the accuracy of the proposed model;
Section IV discusses the experimental results;
and Section V concludes the paper.
\section{METHODS}
\label{sec:FEM}
From~\cite{Niemz2019}, the thermal dynamics of laser-irradiated tissue 
are governed by a partial differential equation of the form
\begin{equation}
    c_v\frac{\partial T}{\partial t} =  
    \nabla \cdot(\kappa\nabla T) + S,
    \label{eq:heat-eq}
\end{equation}
where $T$ is the tissue temperature, $t$ denotes time, and
$c_v$ and $\kappa$ are two tissue-specific physical parameters,
i.e., the \textit{volumetric heat capacity},
and the \textit{thermal conductivity}, respectively.
Readers familiar with heat transfer theory will recognize
Eq.~\eqref{eq:heat-eq} as the well-known \textit{heat equation}
used to describe heat conduction in solids, with the addition of 
an input term $S$, which denotes the heating produced by the laser.
In first approximation, this term can be calculated as~\cite{Niemz2019}
\begin{equation}
    S = \mu_a I,
    \label{eq:heat-generation}
\end{equation}
where $\mu_a$ is the \textit{coefficient of absorption} of the tissue and $I$ is the intensity of the laser beam.
More accurate models for $S$ 
consider light scattering
and other nonlinear optical phenomena and 
are traditionally implemented via Monte Carlo methods~\cite{Niemz2019}.
When light absorption dominates over other optical phenomena, however,
Eq.~\eqref{eq:heat-generation} provides convenient,
computationally inexpensive approximations.
In the following sections, we use the Finite Element Method (FEM) to 
build a model capable of producing numerical solutions to Eq.~\eqref{eq:heat-eq}.
Following the approach outlined in~\cite{Fish2007},
we begin by establishing boundary conditions in order to 
obtain the \textit{strong form} of the problem. 
We then proceed with the derivation of the \textit{weak
form}. Finally, we divide the problem domain 
into discrete elements and apply the Galerkin method
to construct candidate solutions over each element.
The end result of our modeling is a set of algebraic
equations that can be solved numerically
to approximate solutions to the original differential
equation.
\subsection{Derivation of the Strong Form}
\begin{figure}
    \centering
    \includegraphics[width=1.0\linewidth]{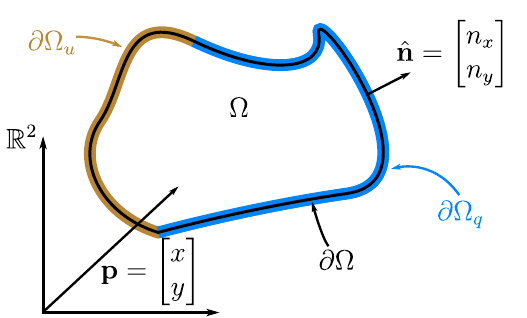}
    \caption{Graphical representation of an arbitrary two-dimensional (2D) domain $\Omega\subset \mathbb{R}^2$.
    A point within this domain is represented by $\mathbf{p} = (x,y)$.
    We define the boundary (a closed line, in this 2D example) as $\partial \Omega$, with local 
    unit normal $\hat{\mathbf{n}}$. The boundary is partitioned into the Dirichlet boundary, $\partial \Omega_u$, and the Neumann boundary, $\partial \Omega_q$. The two boundaries do not intersect but span the entire boundary,
    i.e., $\partial \Omega_u \cup \partial \Omega_q = \partial \Omega$.}
    \label{fig:Domain}
\end{figure}

For the sake of exposition, it is convenient to rewrite 
Eq.~\eqref{eq:heat-eq} into the following equivalent form:
\begin{equation}
    c_v \frac{\partial u}{\partial t} = \nabla \cdot (\kappa \nabla u) + f(\mathbf{p})~\text{in}~\Omega\times[0,t_f],
    \label{eq:strong-form}
\end{equation}
where $u$ is an unknown function of space and time,
$\Omega \subset \mathbb{R}^3$ is an arbitrarily shaped, three-dimensional
spatial domain over which $u$ is defined, $\mathbf{p} = (x, y, z)$
denotes a location within the domain, $[0,t_f]$ denotes
the temporal domain, and $f(\mathbf{p})$ is an arbitrary 
integrable function, equivalent to the input term $S$ in Eq.~\eqref{eq:heat-eq}.
To solve Eq.~\eqref{eq:strong-form}, we prescribe the following
boundary conditions:
\begin{align}
    u = u_g \text{ on } \partial \Omega_u \label{eq:DBC}\\ 
    (\kappa\nabla u) \cdot \hat{\mathbf{n}} = q_n \text{ on } \partial\Omega_q,
    \label{eq:NBC}
\end{align}
as well as the initial conditions
\begin{equation}
    u(\mathbf{p},0) = u_0.
    \label{eq:initial-condition}
\end{equation}
Equations~\eqref{eq:DBC} and~\eqref{eq:NBC} describe Dirichlet and
Neumann boundary conditions, respectively, on the spatial domain $\Omega$.
As illustrated in Fig.~\ref{fig:Domain}, the two boundaries $\Omega_u$ and
$\Omega_q$ do not intersect, and we further assume that their union
encapsulates the entire domain.
Within the context of heat transfer, Dirichlet boundaries are used to prescribe the temperature at a boundary, i.e., to create heat sinks.
Neumann boundary conditions are instead suitable to prescribe the flux (i.e., the heat exchange) at a boundary. In Eq.~\eqref{eq:NBC}, $\kappa$ is the same thermal conductivity term introduced earlier, and $\hat{\mathbf{n}}$ is a unit vector normal to the boundary (refer to Fig.~\ref{fig:Domain}).
As we shall see later in Section~\ref{sec:bound-cond}, our proposed FEM model uses Neumann boundary conditions to describe the heat convection that occurs on the tissue surface exposed to air and Dirichlet boundary conditions to model heat transfer where contacts occur.
\subsection{Derivation of the Weak Form}
To derive the weak form of the problem, let us introduce an unknown,
smooth function $w$ such
that $w = 0$ on the Dirichlet boundary $\partial \Omega_u$.
In the FEM literature, $w$ is referred to as
a \textit{weighting} function.
Multiplying both sides of Eq.~\eqref{eq:strong-form} 
by $w$ and
integrating over the domain, we obtain
\begin{equation}
    \int_{\Omega} wc_v \frac{\partial u}{\partial t}dV = \int_{\Omega}w \nabla \cdot (\kappa \nabla u)dV + \int_{\Omega} w f(\mathbf{p}) dV,
    \label{eq:weak-form}
\end{equation}
where $dV$ represents an infinitesimally small 
volume within $\Omega$.
To ensure convergence of the integrals in Eq.~\eqref{eq:weak-form}, we require both $w$ and
$u$ to have square-integrable first derivatives over $\Omega$.
The equation above can be further rewritten by applying the divergence theorem and Green's formula~\cite{Fish2007}, yielding
\begin{multline}
    \int_{\Omega} wc_v \frac{\partial u}{\partial t}dV + \int_{\Omega} (\nabla w)\cdot (\kappa \nabla u) dV =\\ + \int_{\Omega} w f(\mathbf{p}) dV + 
    \int_{\partial \Omega_q} wq_ndS.
    \label{eq:div-theorem}
\end{multline}
Compared to Eq.~\eqref{eq:weak-form}, this new
relation only contains first-order 
derivatives. 
Note that the last element on the right-hand side is a surface integral across $\partial \Omega_q$, i.e., the boundary on which we have imposed Neumann boundary conditions, with $dS$ representing an infinitesimal area on such surface.
Throughout the remainder of this section, we illustrate 
how to numerically construct functions $w$ and $u$ that
satisfy Eq.~\eqref{eq:div-theorem}.
\subsection{Domain Discretization and Construction of the Solutions}
\begin{figure}
    \centering
    \includegraphics[width=\linewidth]{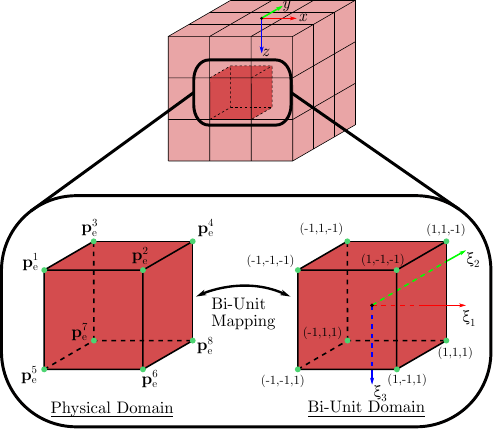}
    \caption{Domain discretization and bi-unit domain. Without loss of generality,
    here we assume tissue specimens to be shaped in the form of a cuboid,
    with a left-handed global frame on the top surface.
    The tissue geometry is partitioned into an arbitrary number $N_{el}$ of
    cuboid-shaped \textit{elements}, each equipped with eight \textit{nodes}.
    The coordinates of each node with respect to the global frame are 
    denoted with $\mathbf{p}_e^A$, with the superscript $A$ identifying 
    a specific node ($A = \{1,2,\ldots,8\}$).
    In the FEM method, candidate solutions for each element are built within a
    \textit{bi-unit} domain, where local
    coordinates are bounded between
    $-1$ and $1$ along each axis. 
    Eq.~\eqref{eq:domain-mapping} provides a mapping between the physical domain and such 
    bi-unit domain.
    }
    \label{fig:elementNodes}
\end{figure}

Let us partition the spatial domain $\Omega$ into
an arbitrary number of discrete subdomains (i.e., \textit{elements})
$\Omega^e$, where $e = 1,...,N_{el}$.
While the shape and structure of the subdomains may be arbitrary, in this paper we shall consider cuboid elements with eight \textit{nodes} per element, as illustrated in Fig.~\ref{fig:elementNodes}.
We define $u_e$ and $w_e$ as 
local approximations of $u$ and $w$ over $\Omega^e$.
With these terms, Eq.~\eqref{eq:div-theorem} can be
rewritten by summing the contributions from each element:
\begin{multline}
    \sum^{N_{el}}_{e=1} \left[\int_{\Omega^e} w_e c_v \frac{\partial u_e}{\partial t}dV + \int_{\Omega^e} (\nabla w_e)\cdot (\kappa \nabla u_e)dV\right] = \\ \sum^{N_{el}}_{e=1} \left[\int_{\Omega^e} w_e f(\mathbf{p}) dV +  \int_{\partial \Omega_q^e} w_e q_n dS\right].
    \label{eq:finite-weak-form}
\end{multline}
In FEM, $u_e$ and $w_e$ are built within a 
\textit{bi-unit} domain (refer to Fig.~\ref{fig:elementNodes}), where
a set of coordinates is represented by
$\boldsymbol{\xi} = (\xi_1, \xi_2, \xi_3)$
and coordinates are bounded between $-1$ and $1$
along each axis.
The mapping between the bi-unit domain and the physical domain
is given by
\begin{equation}
    \mathbf{p}_e(\boldsymbol{\xi}) =  \sum_{A=1}^{N_{ne}} N^A(\boldsymbol{\xi})\mathbf{p}_e^A,
    \label{eq:domain-mapping}
\end{equation}
where $N_{ne} = 8$ is the number of nodes in the element and $N^A(\boldsymbol{\xi})$ are \textit{trilinear} shape functions, i.e., 
\begin{equation}
    N^A(\boldsymbol{\xi}) = \frac{1}{8}(1 + \xi_1^A\xi_1)(1+\xi_2^A\xi_2)(1 + \xi_3^A\xi_3),
    \label{eq:shape-function}
\end{equation}
where $(\xi_1^A, \xi_2^A , \xi_3^A)$ is the location of node $A$. 
With these definitions, we can finally construct candidate
solutions $u_e$ and $w_e$ as
\begin{equation}
    u_e = \sum_{A=1}^{N_{ne}} N^A(\boldsymbol{\xi})d^A_e = \mathbf{N}^T\mathbf{d}_e
    \label{eq:u-he}
\end{equation}
\begin{equation}
    w_e = \sum_{A=1}^{N_{ne}} N^A(\boldsymbol{\xi})c^A_e = \mathbf{N}^T\mathbf{c}_e,
    \label{eq:w-he}
\end{equation}
where $d^A_e$ and $c^A_e$ are nodal degrees of freedom. 
To approximate the temperature across the domain, it 
is necessary to calculate $\mathbf{d}_e$ for
every element, as we show in the following section.
\subsection{Matrix-Vector Formulation}
Using Eqs.~\eqref{eq:u-he} and \eqref{eq:w-he}, it is possible to obtain a more compact formulation for Eq.~\eqref{eq:finite-weak-form} which also is amenable to numerical implementation.
Let us begin by considering the solutions $u_e$ and
$w_e$ over a single domain element $\Omega^e$.
Eq.~\eqref{eq:div-theorem} can be rewritten locally as
\begin{multline}
    \int_{\Omega^e} w_e c_v \frac{\partial u_e}{\partial t}dV + \int_{\Omega^e} (\nabla w_e)\cdot (\kappa \nabla u_e)dV = \\ \int_{\Omega^e} w_e f(\mathbf{p}) dV +  \int_{\partial \Omega_q^e} w_eq_ndS.
\end{multline}
Substituting Eqs.~\eqref{eq:u-he} and \eqref{eq:w-he}
into the relation above, we obtain
\begin{multline}
    \int_{\Omega^e} \mathbf{c}_e^T\mathbf{N} c_v \mathbf{N}^T\dot{\mathbf{d}}_edV + 
    \int_{\Omega^e} (\nabla \mathbf{N}^T\mathbf{c}_e)^T (\kappa \nabla \mathbf{N}^T\mathbf{d}_e)dV = \\
    \int_{\Omega^e} \mathbf{c}_e^T\mathbf{N} f(\mathbf{p}) dV +  \int_{\partial \Omega_q^e} \mathbf{c}_e^T\mathbf{N}q_ndS.
    \label{eq:elem-int}
\end{multline}
Note that the spatial derivatives and the integrals in
Eq.~\eqref{eq:elem-int} are expressed with respect to
the global frame, whereas the shape functions used to
construct $u_e$ and $w_e$
(i.e., Eq.~\eqref{eq:shape-function}) were defined 
within the bi-unit domain.
To correctly apply $\nabla$ to the shape functions,
we need to apply the chain rule:
\begin{equation}
    \nabla = J^{-T}\nabla_{\boldsymbol{\xi}} ,
\end{equation}
where
$\nabla_{\boldsymbol{\xi}} = \left( \frac{\partial }{\partial \xi_1}, \frac{\partial }{\partial \xi_2},\frac{\partial }{\partial \xi_3}\right)^T$ 
and $J = \frac{\partial \mathbf{p}}{\partial\boldsymbol{\xi}}$ denotes the
Jacobian matrix.
Analogously, we change the bounds of integration from the element cuboid in the global reference frame to the cuboid in the bi-unit domain, where $dV = |J|dV^{\boldsymbol{\xi}}$.
This allows us to rewrite Eq.~\eqref{eq:elem-int} in a matrix-vector form:
\begin{multline}
    \mathbf{c}^T_e  
     \underbrace{\int_{\Omega^e_{\boldsymbol{\xi}}} \mathbf{N} c_v \mathbf{N}^T |J|dV^{\boldsymbol{\xi}}}_{\mathbf{M}_e}~
     \dot{\mathbf{d}}_e + \\
    \mathbf{c}^T_e
    \underbrace{\int_{\Omega^e_{\boldsymbol{\xi}}}(J^{-T}\nabla_{\boldsymbol{\xi}}\mathbf{N}^T )^T(\kappa J^{-T}\nabla_{\boldsymbol{\xi}}\mathbf{N}^T) |J|dV^{\boldsymbol{\xi}}}_{\mathbf{K}_e}~
    \mathbf{d}_e = \\
    \mathbf{c}^T_e
    \underbrace{\int_{\Omega^e_{\boldsymbol{\xi}}} \mathbf{N}\mathbf{N}^T \mathbf{f}_e |J|dV^{\boldsymbol{\xi}}}_{\mathbf{F}^{int}_e} +
    \mathbf{c}^T_e 
    \underbrace{\int_{\partial \Omega^e_{q,\boldsymbol{\xi}}} \mathbf{N}q_n|J|dV^{\boldsymbol{\xi}}}_{\mathbf{F}^q_{e}}.
    \label{eq:el-vec-mat}
\end{multline}
Note that we approximated $f(\mathbf{p})$ using the same trilinear shape functions used
earlier to build $u_e$ and $w_e$,
with nodal degrees of freedom $f^1_e, ..., f^8_e$. 
We now aggregate local solutions to assemble global matrices:
\begin{equation}
\sum^{N_{el}}_{e=1} (\mathbf{c}^T_e\mathbf{M}_e\dot{\mathbf{d}}_e + \mathbf{c}^T_e \mathbf{K}_e \mathbf{d}_e) = 
\sum^{N_{el}}_{e=1} (\mathbf{c}^T_e\mathbf{F}^{int}_e + \mathbf{c}^T_e\mathbf{F}^q_{e}),
\end{equation}
which can be more compactly rewritten as
\begin{equation}
    \mathbf{M}\dot{\mathbf{d}} + \mathbf{K} \mathbf{d} = 
 \mathbf{F}.
\label{eq:matrix-vec-form}
\end{equation}
In the equation above, $\mathbf{M}$ is the thermal mass matrix,
$\mathbf{K}$ is the thermal conductance matrix,
$\mathbf{F}$ is the heat source vector,
and $\mathbf{d}$ is the tissue temperature at each node in the mesh. 
Note that Eq.~\eqref{eq:matrix-vec-form} is equivalent to Eq.~\eqref{eq:finite-weak-form} but written as a linear ordinary differential equation.
By solving Eq.~\eqref{eq:matrix-vec-form} for $\mathbf{d}$, we can determine the temperature at any location in the mesh.
\subsection{Time Stepping}
\label{sec:time-stepping}
Equation~\eqref{eq:matrix-vec-form} is a first-order differential equation that can be solved with discrete time stepping.
Let us denote
$\mathbf{d}_n$ and $\mathbf{v}_n$ to be the discrete approximations of $\mathbf{d}(t_n)$ and $\dot{\mathbf{d}}(t_n)$ respectively.
This allows us to rewrite Eq.~\eqref{eq:matrix-vec-form} as
\begin{equation}
    \mathbf{M}\mathbf{v}_{n+1} + \mathbf{K}\mathbf{d_{n+1}} = \mathbf{F}_{n+1}.
    \label{eq:discrete-ode}
\end{equation}
Using the Crank-Nicolson method, we can define $\mathbf{d}_{n+1}$ as 
\begin{align}
    \mathbf{d}_{n+1} = \mathbf{d}_n + \frac{\Delta t}{2}(\mathbf{v}_{n+1} + \mathbf{v}_n),
    \label{eq:crank-nicolson}
\end{align}
where $\Delta t$ is the duration of the time step.
Substituting Eq.~\eqref{eq:crank-nicolson} into Eq.~\eqref{eq:discrete-ode} and solving for $\mathbf{v}_{n+1}$ we get
\begin{equation}
    \mathbf{v}_{n+1} = \left(\mathbf{M} +  \frac{\Delta t}{2} \mathbf{K}\right)^{-1 }\left(\mathbf{F}_{n+1} - \mathbf{K}(\mathbf{d}_{n} + \frac{\Delta t}{2}\mathbf{v}_n)\right).
\end{equation}
This expression defines $\mathbf{v}_{n+1}$ as only a function of $\mathbf{d}$ and $\mathbf{v}$ at the previous time step, $n$.
We initialize $\mathbf{d}_0$ with the starting temperature of the mesh and $\mathbf{v}_0$ using Eq.~\eqref{eq:discrete-ode}.
Then at each time step $n$, we first solve for $\mathbf{v}_{n+1}$ and then for $\mathbf{d_{n+1}}$. 
\subsection{Boundary Conditions}
\label{sec:bound-cond}
In our model, we allow each element on the external
surface of the domain to have one of three
boundary conditions: a heat sink, constant flux,
or convection boundary. 
In the case of a heat sink, we apply Dirichlet boundary conditions to the surface, restricting the temperature on the surface to some fixed value.
For constant flux conditions, a Neumann boundary is
applied to the surface and $q_n$ is set to the flux
value.
A special case of the flux boundary is a convection
boundary, which models the heat transfer created by
a fluid passing over a solid. 
We model such a flux using Newton's Law of
Cooling~\cite{Osullivan1990}, i.e.,
\begin{equation}
    q_n = h(T_\infty - u),
    \label{eq:convection}
\end{equation}
where $T_\infty$ is the temperature of the fluid and $h$
is the heat transfer coefficient.
\section{EXPERIMENTAL VERIFICATION}
\subsection{Experimental Setup}
\begin{figure}
    \centering
    \includegraphics[width=\linewidth]{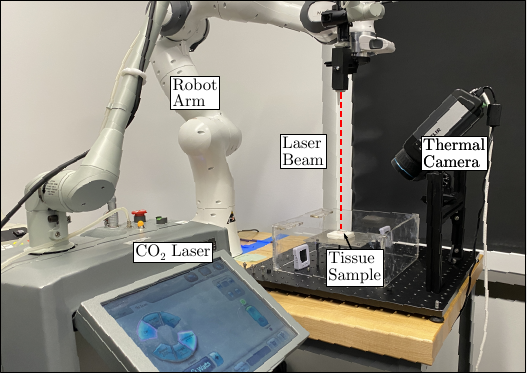}
    \caption{Experiments used a surgical CO$_\text{2}$ laser  
    whose beam is delivered through an articulated
    (passive) arm.
    The tissue surface temperature was monitored with an infrared thermal camera
    at a rate of 20 frames per second (fps),
    and spatial resolution of 70~$\frac{\text{pixel}}{\text{cm}}$.
    The distance between the beam's focal point and the tissue surface ($d_f$)
    was controlled with a robotic arm.}
    \label{fig:exp-setup}
\end{figure}

To verify the accuracy of the proposed FEM model, we conducted 
laser-tissue interaction experiments using the setup shown in
Fig.~\ref{fig:exp-setup}.
A surgical carbon dioxide (CO$_\text{2}$) laser, the Lumenis Acupulse
(Lumenis, Yokneam, Israel) was used to irradiate soft tissue targets. 
The tissue surface temperature was recorded with an infrared thermal camera,
the A655sc (Teledyne FLIR, Oregon, USA), and compared to the prediction
generated by the FEM model.
In each experiment, the laser was applied for 15 seconds, and the tissue temperature 
was recorded for an additional 15 seconds in order to document cooling.
Two types of tissue targets were used in these experiments, namely \textit{ex-vivo} chicken muscle,
which was sourced from a local butcher shop; and agar-based gelatin, a soft tissue surrogate 
frequently used in thermal laser-tissue interaction studies~\cite{Rastegar1988}.
The agar gels were fabricated in our laboratory using a mixture
of 2\% agar powder (Sigma-Aldrich Chemie, Germany)
and 98\% deionized water.
To further ascertain the generality of our model,
we performed experiments with different levels of 
laser beam focusing.
Intuitively, focusing the laser beam into a tighter spot will increase the beam intensity $I$, and it is expected to produce a stronger thermal response (refer to Eqs.~\eqref{eq:heat-eq} and \eqref{eq:heat-generation}).
In our experimental setup, the laser beam width $w$ is
controlled by regulating the distance $d_f$ between
the laser beam's focal point and the tissue
surface (see Fig.~\ref{fig:exp-setup}).
The relation between
$d_f$ and $w$ can be derived from simple laser optics~\cite{Lee2022}:
\begin{equation}
    w(z) = w_0\sqrt{1 + \left(\frac{\lambda (d_f + z)}{\pi w_0^2}\right)^2},
\end{equation}
where $z$ is the optical axis of the beam, 
$\lambda$ is the laser wavelength,
and $w_0$ is the\textit{ beam waist}
(i.e., the radial width measured at the focal point).
We conducted experiments with $d_f = \{25, 30, 35\}$ \unit{cm}.
\subsection{FEM Simulation Setup}
The FEM model described in Section~\ref{sec:FEM} was implemented in C++ and compiled into a MEX file so that it could be run within the MATLAB environment (The MathWorks, Inc., Natick, MA, USA).
The finite element mesh was initialized as a $(34\times34\times50)$ cuboid, representing a
$(2\times2\times0.5)$ \unit{\centi\meter\cubed} volume.
The initial temperature of the cuboid was initialized to match the initial surface
tissue temperature observed experimentally.
\subsubsection{Boundary Conditions}
\label{sec:boundary-conditions-simulation}
To model the heat exchange between the tissue specimen
and the surrounding environment during each experiment, the FEM
model was configured to use Dirichlet and Neumann boundary conditions.
Specifically, a Dirichlet boundary was imposed on the bottom surface
of the cuboid, to model a heat sink, representing contact between the 
tissue specimen and the underlying experimental bench.
All remaining surfaces were treated as Neumann (i.e., convection)
boundaries, given that they were fully exposed to air.
We adopted a \textit{natural convection} model~\cite{Osullivan1990},
which involves using Eq.~\eqref{eq:convection} and scaling the heat transfer
coefficient $h$ by a value of $(T - T_\infty)^{1/4}$.
Here, $T_\infty$ denotes ambient
temperature, which 
was measured with thermometers
placed around the experimental setup. 
Observed values are reported in Table~\ref{tab:tissue-params}.
\subsubsection{Tissue Physical Properties}
In addition to the initial and boundary 
conditions, our FEM model requires knowledge
of the tissue's thermal and optical
properties.
Table~\ref{tab:tissue-params} lists the
parameters that were used by the simulator and their physical
units.
\begin{table}[]
\centering
    \caption{FEM simulation parameters}
\begin{tabular}{@{}lcc@{}}
\toprule
\textbf{Parameter}            & \multicolumn{1}{l}{\textbf{Agar}} & \multicolumn{1}{l}{\textbf{Chicken}} \\ \midrule
$\mu_a$ (\unit{\centi\metre}) & 31                                & 26                                   \\
$c_v$ (\unit{\joule\per\centi\meter\cubed\per\degreeCelsius})                        & 4.3                               & 3.73                                 \\
$\kappa$ (\unit{\watt\per\centi\meter\per\degreeCelsius})                    & 0.0062                            & 0.0049                               \\
$h$  (\unit{\watt\per\centi\meter\squared\per\degreeCelsius})                         & 0.022                             & 0.029                                \\
$T_\infty$  (\unit{\degreeCelsius})                    & 24                                & 24                                   \\ \bottomrule
\end{tabular}
   \label{tab:tissue-params}
\end{table}
The volumetric heat capacity and the thermal conductivity of the agar phantoms were calculated using the following empirical approximations from~\cite{Niemz2019}:
\begin{align}
    c_v &= (1.55 + 2.8 w)\rho \\
    k &= 0.0006 + 0.0057 w, 
\end{align}
where $w=0.98$ is the water content and
$\rho = 1.00$~\unit{\gram\per\centi\meter\cubed}
is the material density.
For the chicken muscle specimens, we assumed physical
properties similar to that of human muscle
tissue~\cite{ITISDatabase}. The absorption coefficient
$\mu_a$ and heat transfer $h$ can be highly variable,
thus making it impractical to use tabulated values
from prior literature.
Our approach for the selection of these two parameters
was to manually tune them to reduce modeling error.
\subsubsection{Heat Source}
The laser intensity $I$ was simulated based
on the Lambert-Beer law~\cite{Lee2022}:
\begin{equation}
    I = \frac{2P}{\pi w(z)}e^{\frac{-2}{w(z)}(x^2 +y^2) - \mu_a z}.
\end{equation}
Here, $P$ is the laser power, $w$ is the beam's radial width (which can be calculated based on Eq. (24)), and $x$, $y$, and $z$ are the coordinates of a Cartesian reference frame established on the tissue surface,
whose $z$-axis corresponds to the optical axis of the 
laser beam.
\subsection{Results}
\begin{figure*}
    \centering
    \includegraphics{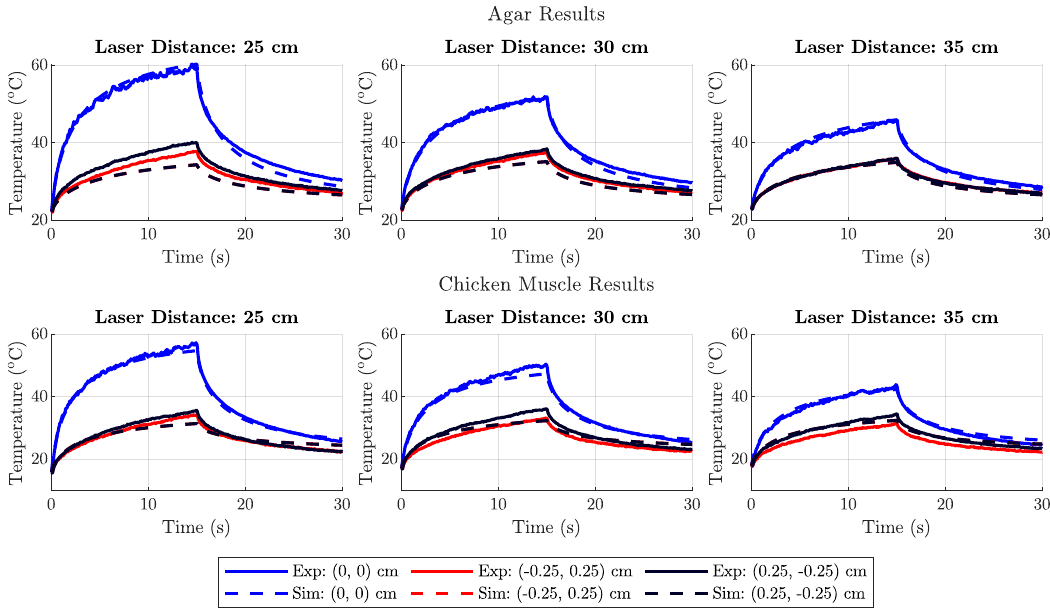}
    \caption{Simulated temperature results and experimental temperature results on agar tissue phantoms (top) and chicken muscle (bottom). The experimental results are the averages from the 5 trials for each combination of laser distance and tissue type.}
    \label{fig:agarResults}
\end{figure*}
Each experimental condition (two tissue types, three 
laser beam focusing levels) was replicated five times,
for a total of 30 experimental runs.
Results are shown in Fig.~\ref{fig:agarResults}.
Temperature profiles are shown for three locations on
the tissue surface, i.e., the incidence point of the
laser at
$(0,0,0)$~\unit{\centi\meter}
and two locations located symmetrically around the
incidence point at \textit{x-} and \textit{y-}coordinates
$(-0.25,0.25)$~\unit{\centi\meter} and
$(0.25,-0.25)$~\unit{\centi\meter}. 
Average temperature tracking errors (root-mean-square error, RMSE) for each experimental condition are reported in 
Tables~\ref{tab:agarResults} and~\ref{tab:chickResults}.
In most experimental conditions, the FEM model predicted the tissue temperature with a tracking accuracy 
within 2~\unit{\degreeCelsius}.
The largest observed RMSE was 3.31~\unit{\degreeCelsius}.
\begin{table}[]
\centering
\caption{Average Temperature RMSE (and Standard Deviation) for Experiments on Agar Specimens}
\begin{tabular}{@{}rlll@{}}
\toprule
\textbf{Focal Distance} $d_f$    & \multicolumn{1}{c}{\textbf{25 \unit{\centi\metre}}} & \multicolumn{1}{c}{\textbf{30 \unit{\centi\metre}}} & \multicolumn{1}{c}{\textbf{35 \unit{\centi\metre}}} \\ \midrule
\textbf{Incidence Point}  & 1.67 (0.31)                     & 1.15 (0.19)                     & 0.95 (0.12)                     \\
\textbf{(-0.25, 0.25) cm} & 1.80 (0.68)                     & 1.27 (0.59)                     & 0.67 (0.10)                     \\
\textbf{(0.25, -0.25) cm} & 3.31 (0.64)                     & 1.80 (0.55)                     & 0.63 (0.12)      \\ \bottomrule              
\end{tabular}
\label{tab:agarResults}
\end{table}

\begin{table}
\centering
\caption{Average Temperature RMSE (and Standard Deviation) for Experiments on Chicken Specimens}
\begin{tabular}{@{}rlll@{}}
\toprule
\textbf{Focal Distance} $d_f$    & \multicolumn{1}{c}{\textbf{25 \unit{\centi\metre}}} & \multicolumn{1}{c}{\textbf{30 \unit{\centi\metre}}} & \multicolumn{1}{c}{\textbf{35 \unit{\centi\metre}}} \\ \midrule
\textbf{Incidence Point}  & 1.53 (0.28)                     & 1.62 (0.37)                     & 1.27 (0.55)                     \\
\textbf{(-0.25, 0.25) cm} & 2.15 (0.69)                     & 1.59 (0.61)                     & 2.18 (0.82)                     \\
\textbf{(0.25, -0.25) cm} & 2.00 (0.81)                     & 1.98 (0.30)                     & 1.49 (0.74)   \\ \bottomrule                 
\end{tabular}
\label{tab:chickResults}
\end{table}
\section{DISCUSSION}
Experimental results show that the proposed FEM model can accurately predict the temperature of laser-irradiated tissue.
This work addresses a gap in the current ecosystem of surgical
robot simulators, laying the foundation for a new physics engine
which will enable the integration of surgical lasers.
Our model was validated through laser experiments on
laboratory-made tissue phantoms and ex-vivo chicken
muscle, achieving overall good tracking accuracy
(see Fig.~\ref{fig:agarResults}).
While these results are promising, further work is
needed to enhance the FEM model and integrate it
in surgical robot simulators.
While the model was verified on ex-vivo tissue,
in-vivo tissue experiences additional cooling effects from perfusion (i.e., blood flow)~\cite{Niemz2019}.
Extending the model to account for perfusion may be
necessary for accurate modeling in tissues with
significant blood flow.
Additionally, our model currently does not predict
thermal tissue damage caused by laser heating.
Incorporating the Arrhenius model~\cite{Niemz2019}
to calculate cellular death would enable the
prediction of coagulation and other physical
processes secondary to the 
heating.
Another limitation of the present model is that 
it was only verified on tissue specimens having
simple geometrical shapes.
In the future, we plan to explore the generation of
the tissue geometry for the FEM simulator based
on medical imaging, analogously to the way in
which Computer-Aided Design (CAD) software can
generate meshes for Finite
Element Analysis of complex parts.
Finally, an analysis of the model's
computational complexity should be performed to
evaluate its efficiency.
In general, the model produced more accurate predictions at the 
laser incidence point rather than surrounding areas.
The most pronounced loss of accuracy was observed in the experiments
with $d_f = 25$ \unit{\centi\metre} (observe the two leftmost plots in Fig.~\ref{fig:agarResults}),
where the laser spot was the tightest and thus produced the strongest
thermal responses.
This loss of accuracy may be due to the way convection cooling was implemented.
Recall from section~\ref{sec:boundary-conditions-simulation} that the simulation used a natural convection model, which
scales the heat transfer coefficient $h$ by the
difference between tissue and ambient temperature. While
we take the temperature at a single location (i.e., the incidence point) to scale the
heat transfer coefficient, the new value applies
uniformly to the entire tissue surface.
Therefore, cooler locations experience a stronger dissipation
effect than they would if the heat transfer coefficient
was scaled uniquely for each location.
This effect could be mitigated by adjusting the
simulation to have a unique heat transfer coefficient for
every element in the mesh. 
Additional errors may arise from the modeling
assumptions made throughout the manuscript.
While we consider convective heat transfer,
our model does not account for \textit{radiative} heat
transfer, which occurs at a rate proportional to the 
fourth power of temperature~\cite{Osullivan1990}.
At higher temperatures, radiation may significantly
cool the tissue. 
Furthermore, a tissue's thermal and
optical properties may vary during laser exposure due to
temperature variations~\cite{Niemz2019},
but these values were held
constant during the simulation. 
Lastly, our model 
neglects scattering and other nonlinear
optical effects. These assumptions may be acceptable in first
approximation~\cite{Niemz2019} but may
contribute to errors.
\section{CONCLUSION}
This paper presented a Finite Element Method (FEM)
model to simulate the thermal tissue
response produced by surgical lasers.
The model was validated through benchtop experiments
on laboratory-made tissue phantoms and ex-vivo
chicken muscle, achieving an RMSE generally 
smaller than 2 \unit{\degreeCelsius}.
These promising results lay the groundwork
for the integration of surgical lasers into
surgical robot simulators, which are currently 
not supported due to the lack of suitable physics 
engines.
Further work is necessary to enhance the 
proposed model's applicability to real surgical
scenarios. Specifically, extending the model to account for perfusion effects in in-vivo tissues is crucial for accurate thermal predictions, as blood flow can significantly influence tissue cooling. Additionally, incorporating the Arrhenius model to predict thermal tissue damage would enable the simulation of coagulation and other temperature-induced physical processes.


\bibliographystyle{IEEEtran}
\bibliography{references.bib}
\end{document}